# A Survey of Plagiarism Detection Systems: Case of Use with English, French and Arabic Languages

Mehdi Abdelhamid[1], Sofiane Batata[1] and Faiçal Azouaou[1]

[1]*Ecole nationale Supérieure d'Informatique, ESI, Oued Smar, Algiers, Algeria ([https://www.esi.dz](https://www.esi.dz))*
*Email address:* (gm_abdelhamid, s_batata, f_azouaou)@esi.dz

**Abstract:** In academia, plagiarism is certainly not an emerging concern, but it became of a greater magnitude with the popularisation of the Internet and the ease of access to a worldwide source of content, rendering human-only intervention insufficient. Despite that, plagiarism is far from being an unaddressed problem, as computer-assisted plagiarism detection is currently an active area of research that falls within the field of Information Retrieval (IR) and Natural Language Processing (NLP). Many software solutions emerged to help fulfil this task, and this paper presents an overview of plagiarism detection systems for use in Arabic, French, and English academic and educational settings. The comparison was held between eight systems and was performed with respect to their features, usability, technical aspects, as well as their performance in detecting three levels of obfuscation from different sources: verbatim, paraphrase, and cross-language plagiarism. An in-depth examination of technical forms of plagiarism was also performed in the context of this study. In addition, a survey of plagiarism typologies and classifications proposed by different authors is provided.

**Keywords:** plagiarism detection, plagiarism detection systems, software comparison, plagiarism types, plagiarism obfuscation, technical exploits plagiarism, cross-language plagiarism.

## 1. Introduction

In academia and research, plagiarism is one of the most prominent forms of professional dishonesty and misconduct. The definition given by Teddi Fishman (2009), director of the International Center for Academic Integrity, was able to delimit plagiarism in an accurate manner, according to which, plagiarism occurs when the author of a work that is expected to be original regardless of its type or format, uses words, ideas, or work products that are attributable to an identifiable source other than the author himself, without attributing that work to said source, all in order to claim a sort of credit or benefit that need not be monetary.

While the fight against plagiarism and its ethical implications are not a novel issue for academia and elsewhere, it remains observable that there are always a few people who are willing to take the unscrupulous shortcut towards academic recognition. Plagiarists are still coming up with unique and creative methods to cheat the system and get by unnoticed, despite the popularisation of automated anti-cheating support programs, and on which reviewers have to depend as the task of exposing plagiarism in any form of work became hardly feasible with sheer human effort. In a fashion similar to fraud detection, teachers and reviewers alike are in the right to question the robustness and usefulness of current plagiarism detection systems when faced with plagiarism instances that go beyond simple copy-paste and even beyond languages barriers, all of which are questions we will attempt to tackle throughout this paper: What novel means are plagiarists employing to efficiently obfuscate their plagiarism attempts? are modern plagiarism detection systems useful against such unusual customs? are they able to handle a base of source content as wide and borderless as what is accessible to any person nowadays or are they still limited in their scope?

Plagiarism is often confused with copyright infringement. While the two are similar in some aspects, either of them can occur without the other.

An example of a copyright infringement case is the act of illegally printing and selling copies of a popular book, but this does not constitute a plagiarism case since the vendor is not claiming to be the author of that book.

On the other hand, a student that copy-pastes from Wikipedia for his homework is indeed a case of plagiarism, but it does not infringe any copyrights since the source is under a creative commons license. It is also worth noting that the simple act of paraphrasing is deemed enough to counter most infringement issues when copyright is due and, in most cases of academic plagiarism, the original authors are not missing on any of their rights, as they can't pretend to the benefits that the plagiarists are reaching for, which usually amounts to an academic degree or a grade on an essay or homework.

In a broader definition, it can be considered that any form of reproduction without attribution in academic works is an instance of plagiarism, but anti-plagiarism policies usually exclude the following cases:

- o Quoted works that fall under the public domain or are reproduced with all necessary permission and/or attribution.
- o References, bibliographies, tables of content, prefaces, and acknowledgements.
- o Generic terms, laws, standard symbols, and equations.
- o Similarities of minor nature, that are more susceptible to be mere coincidences.

In a survey of the state of research on plagiarism detection performed by Foltýnek et al. (2019) the authors propose and discuss a three-layered model to structure the literature on academic plagiarism detection as represented in figure 1. This model distinguishes between three broad aspects for addressing the topic and problem of plagiarism detection.

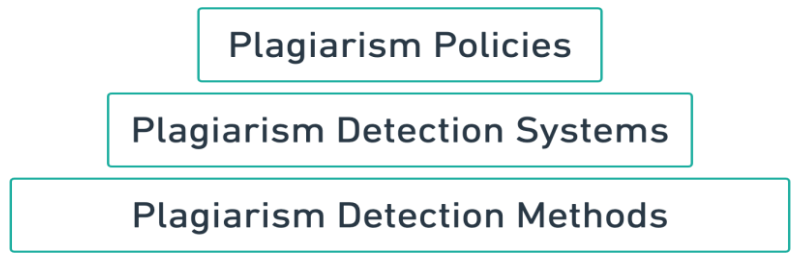

**Figure 1: Three-layered model of plagiarism detection** (Foltýnek et al., 2019)

- o **Plagiarism policies**: the systems and processes that can be set up by universities and institutes to minimize the prevalence of plagiarism, by establishing tolerance thresholds as well as the repercussions and protocols to follow when a plagiarism case has been recognized and the workflow leading towards that.
- o **Plagiarism detection systems**: software that implement the detection methods in order to take documents as input, process them and output an assessment of originality, usually represented as a measure of similarity in percentage, as well as retrieved sources.
- o **Plagiarism detection methods**: the algorithms built to detect similarity on a file against a corpus of sources. Most methods differ in terms of the level of language on which they perform: lexical, syntactic or semantic. Some other methods make use of non-textual features like mathematical formulas, references or images.

In this paper, our interest is accordance with the second layer, that of the plagiarism detection systems (PDS).

We define a **plagiarism detection system** as *software that offers the necessary data and information to support a human reviewer in asserting whether a document, or parts of it, are an instance of plagiarism, as well as the severity of the case and the sources from which it was copied.*

Thus, it is currently expected from that type of software to output a report showcasing all the passages identified as being similar or matching to passages in other documents that are made accessible to the reviewer by providing links pointing to them (Weber-Wulff, 2015).

Plagiarism detection is, at present, not an automated task but a computer-assisted work. Identifying this type of software as "plagiarism detectors" is a misnomer, as they do not detect plagiarism per se, but merely point cases of textual similarity from sources. Those cases may or may not be considered as plagiarism, and it is still always up to the reviewer to decide upon that.

More precise terms are used in literature to refer to that type of software like text-matching systems, similarity assessment software, or source detection. In order not to confuse the reader, we will use the term "Plagiarism Detection System", or PDS for short, to speak about said type of software in this paper.

PDS are also actively used for what would appear to be their reverse purpose, by submitting an original work in order to discover illegal reproductions in other sources. This is done more generally for web content like articles, allowing to detect infringement cases and claim copyrights.

In academic settings, we can consider an actual plagiarism detection system to be the combination of the reviewing committee (or board of ethics of the institute), their workflow and processes, plus the software they use to support them with proofs and sources in order to ascertain their decisions regarding the fact that a case in hand is to be deemed as plagiarism.

## 1.1 The importance of addressing academic plagiarism

Fishman states that the harm in plagiarising lies in the fact that the author or student is asking for recognition in doing a work from which no benefit was earned (Fishman, 2009).

In an academic context, it is clear that the severity of plagiarism differs greatly from the case of a student submitting a plagiarised homework to claim a grade, to that of a researcher defending a plagiarised thesis in order to claim a degree.

Despite this gap in severity, it is still important to raise awareness over plagiarism among students as soon as possible. A number of studies (Park, 2003; Comas-Forgas and Sureda-Negre, 2010; Selemani et al., 2018; Shang, 2019) shed the light on the students' behaviour, methods, and reasons behind their acts of plagiarism. Students admit to plagiarising because they lack self-confidence in their capacity to produce valuable work or simply out of laziness. In other cases, that are more common than it would seem, they can even be in ignorance of what is considered plagiarism, what are its consequences, and how they can avoid it. This was demonstrated through the survey conducted by Shang (2019), which revealed that only a minority of students correctly understood what constitutes self-plagiarism, or the unethicality of paraphrasing portions of information without accurate citation. On the other hand, a majority of students surprisingly expressed that plagiarism was not a big deal and reported being tempted to plagiarize because everyone else was doing it (Shang, 2019). All of these reasons validate the necessity to educate and inform students about plagiarism and the scholar's ethics throughout their curriculum.

While students in their early years reach for plagiarism for what eventually amounts to an insignificant gain, the same cannot be said when it comes to fully-fledged research papers and thesis. While these works may or may not be funded, they are still always expected to be original, and their authors are always credited in some form like a degree, a grant, or even just a publication on their name, as that adds to their notoriety in their field. Needless to add that a work that is a product of plagiarism contributes next to nothing to its field or literature. Smart and Gaston (2019) surveyed journal editors from around the world and reported, for journals that check all articles on submission, that 24.8% of all articles showcased a plagiarism score exceeding 10% (with 10.85% of that portion representing articles showing over 25% of similarity). They also noticed that Asian editors experienced the highest levels of plagiarized or duplicated content, compared to their European and North American peers.

In research, plagiarism can also distort meta-studies, which make conclusions based on the number or percentage of papers that confirm or refute a certain phenomenon through their experiments. Thus, if a portion of those papers is plagiarised from other ones, then the number of experiments that were actually conducted is lower than what it was accounted for, and the conclusions presented by that meta-study would be incorrect (Foltýnek et al., 2019).

It becomes clear that, alongside raising awareness on the matter, plagiarism detection is of the utmost importance in research and academia. As the task became hardly feasible through sheer human effort with the advent of ubiquitous information access, it justifies the need for plagiarism detection tools that is expressed by schools, universities, research centers, publishers, and journals.

## 1.2 Limitations of Plagiarism Detection Systems

Automated similarity detection is a computational and development demanding task, it is hard to achieve and maintain for most institutes expressing the need to counter plagiarism, so it becomes interesting for them to outsource the fulfilment of that need by acquiring solutions that are developed, maintained, and sold as a service by third parties. And indeed, many software-service companies emerged around the business of academic plagiarism detection. The systems they put on the market evolved swiftly during the past two decades, but still, they suffer from limitations up to this day:

- the detection quality varies significantly across languages,
- they cannot detect all forms of plagiarism, with a detection rate being anti-proportional to the degree of obfuscation,
- they are helpless against cross-language, or translation plagiarism,
- they can be vulnerable to techniques used by plagiarists that do not even involve textual obfuscation,
- they can be unintuitive for the users, or show results that are hard to interpret,

It is worth noting that plagiarism detection highly favours false positives over false negatives, so it is always up to the user to decide upon each case highlighted by the detection system, this falls in line with our prior statement that those systems do not detect plagiarism per se, but just cases of textual similarity that are to be ascertained by a human reviewer.

### 1.3 What is done in this survey

In this survey, we try to assess the capacity of several plagiarism detection systems to detect plagiarism in three languages: English, Arabic, and French, as well as their performances with different levels of obfuscation. Besides English, the inclusion of both Arabic and French was a response to their under-representation in surveys performed so far.

We also put a special highlight on the detection of translation plagiarism, which is not at all a novel method for plagiarists, but that was still highly undetectable some years ago. Finally, we also investigate advanced obfuscation techniques that are used to dupe detection systems without altering the text content itself.

### 1.4 Organisation of the paper

The remainder of this paper is organised as follows: First, is presented a survey of related works on the classification of plagiarism types and comparison of similarity detection systems, as well as a discussion of the authors' proposed typology. This is followed by a description of the methodology and corpora used to carry out the tests for this study, then a presentation of the systems that were considered in the study. After reporting and discussing the acquired results, recommendations points are given, then a conclusion for this work.

## 2. Survey of Related works
### 2.1 Typologies of plagiarism

Plagiarism can occur in all types of works, including text, source code or audio-visual content. It can be either intended by its author, or accidental, going as far as possibly being merely coincidental. Plagiarism in text is by far the most investigated and spread form.

Researchers proposed a variety of typologies for textual plagiarism, and their classifications differ in regard to the criteria used to distinguish between instances of plagiarism as well as their scope. Some authors use different terms to describe a similar form of plagiarism and, overall, there is a visible overlap between all typologies.

Maurer et al. (2006) provide a broader categorisation of plagiarism based on the intent of its perpetrator. They distinguish between 4 types as described in table 1. Interestingly, they also consider as plagiarism the act of "copying so many words or ideas from a source that it makes up the majority of the work, whether credit is given or not"

**Table 1 Plagiarism typology proposed by Maurer, Kappe and Zaka**

| Plagiarism type | Description |
|---|---|
| Unintentional plagiarism | The same ideas may be worded similarly under the effect of influence or by coincidence. |
| Intentional plagiarism | The deliberate act of copying all or part of someone's work, without properly citing them. |
| Accidental plagiarism | Is more common among students due to their lack of knowledge on the proper ways to cite and paraphrase in order to avoid plagiarism. |
| Self-plagiarism | Occurs when using, without referencing, one's own published or submitted work in another form or situation. |

Eassom (2016) considers not only the ways of copying and reproducing text, but also the cases of research and academic misconduct in which plagiarism is a medium. She thus distinguishes between 10 types of plagiarism in research that are detailed in table 2.

**Table 2 Plagiarism typology proposed by Eassom**

| Plagiarism type | Description |
|---|---|
| Verbatim plagiarism | Copying someone's words without citing them, or by citing them without indicating that it is a direct quote. |
| Complete plagiarism | Taking a whole work or study and submitting it as is under the perpetrator's name. |
| Paraphrasing | Simply rephrasing someone's writing, or completely rewriting it, keeping only the ideas or concepts. |
| Secondary source or inaccurate citation | Referencing the sources that are cited in a work without mentioning the main work itself, providing a false sense of the amount of review performed by the perpetrator. |
| Invalid source or misleading citation | Referencing incorrect or non-existent sources in an attempt to extend the list of references. |
| Duplication | Reusing one's own previous work or study, without attribution. |
| Repetitive research | Replicating text or data from similar works with a similar methodology, without attribution. |
| Replication | Submitting a paper to multiple publishers, resulting in the same work being published more than once. |
| Misleading attribution | Providing an inaccurate list of authors in the work, either by excluding authors who made considerable contribution to the study, or by including others who did not contribute. |
| Unethical collaboration | Using works, results or ideas that are a product of a collaboration, without citing the collaborative nature of the study or the persons involved. |

Weber-Wulff (2014) on the other hand suggested a typology that focuses more on the techniques used to plagiarise, as it considers the modifications applied to the source from which the perpetrator has copied, as well as how he pastes to his work. Her typology is detailed in table 3.

**Table 3: Plagiarism typology proposed by Weber-Wulff**

| Plagiarism type | Description |
|---|---|
| Copy & Paste | Simply taking from the original work without any modification. |
| Shake & Paste Collections | Copy-pasting without modification, but from multiple sources. |
| Disguising the source text | Replacing words and phrases, or modifying the order of sentences but failing to properly cite the source. This is often thought by students to be proper paraphrasing. |
| Clause Quilts, mosaic or patch-writing plagiarism | Taking from one or multiple sources and "quilting" or stitching them with one's own words and phrases. |
| Translation plagiarism | Can be done either manually or by making use of online machine translation services. Retracing a source in a language other than that of the analysed work is still very challenging for detection software. |
| Structural plagiarism | Goes beyond simple paraphrasing the source text, by also copying the source's organisation, arguments and overall structure. The detection of this form is challenging as it is not a simple linguistic copy, but fundamental ideas plagiarism. |
| Pawn sacrifice | One of the only forms where the actual source is cited, but the exact amount of work taken from it is unclear. |
| Cut & Slide | Similar to pawn sacrifice, with the exception that only a portion of the source is appropriately documented, while the rest is downgraded in importance. |
| Self-Plagiarism | The source belongs to the perpetrator, but gets reused in a situation where originality is expected. Includes submitting a paper for publication in a number of journals without a reprint notice, which would effortlessly boost the number of publications under the author's name, thus benefiting multiple times from a single work. |

A white paper published by the plagiarism detection system "Turnitin" (2012) divides the spectrum of plagiarism committed by students into 10 distinct forms, in regard to how the student's work compares to its sources. They are listed as follows, from the most to least prevalent one:

The student's work is a:

- **Clone**: if it is an exact copy of someone else's work,
- **Mash-up**: if it copies content from multiple sources with little to no change,
- **CTRL-C**: if it contains several portions copied from a single source, with no alterations,
- **Remix**: if it copies and paraphrases from multiple sources.
- **Recycle**: if it contains a large portion of work that was previously submitted by that same student (self-plagiarism).
- **Retweet**: if it correctly references the original source but remains close to identical in its wording, without quoting it.
- **Find-Replace**: if it just replaces key words or phrases from the source.
- **Aggregator**: if it does actually reference its sources, but is just a mash-up, meaning that it contains close to no original work.
- **Error 404**: if it references sources that are either non-existent or inaccurate.
- **Hybrid**: if it contains both correctly cited sources, to give the impression of a correct work, and plagiarised passages.

Kumar and Boriwal (2019) Kumar and Boriwal (2019) present another typology that varies both in terms of the plagiarist's intent as well as the methods employed to disguise the act, as reported in table 4.

**Table 4: Plagiarism typology proposed by Kumar and Boriwal**

| Plagiarism type | Description |
|---|---|
| Deliberate plagiarism | The process of actively trying to deceive by passing off someone else's work as their own. |
| Paraphrasing | The act of altering a few words but retaining the same sentence structure used in the original source |
| Patchwork Paraphrasing | Which is paraphrasing, but performed on passages that were taken from multiple sources before stitching them together. |
| Bluffing | By reading key source texts and noting the key ideas so that they seem different though, in essence, they are the same. |
| Stitching sources | When the overall work is mainly composed of multiple sources that are combined together to compose a new work, even if legitimate references are done. |
| Using a copy of own work | By reusing previous works to answer new assignments, taking it solely as a manner to save time and effort. |
| Mosaic plagiarism | By using synonyms to substitute words of existing text in an attempt to pass off the submitted work as original. |
| Accidental plagiarism | This can manifest as a mistaken attribution to another author, a neglected source to be cited, or simple unintentional plagiarism through paraphrasing or patchwork paraphrasing. |
| Buying assignments | A form of ghost-writing that is more common among students, which involves buying assignments from other students or from paper-mills, and submitting as their own individual work. |

Foltýnek, Meuschke, and Gipp (2019) discuss a typology of academic plagiarism with a more technical perspective, by considering the natural language layer that is affected. The forms are ordered increasingly by their level of obfuscation in table 5.

**Table 5: Plagiarism typology proposed by Foltýnek, Meuschke and Gipp**

| Plagiarism type | Description |
|---|---|
| Character-preserving plagiarism | Is limited to direct copying of sources without alteration, and regardless of how the content is pasted to the resulting work. This can even occur while citing the sources, which were detailed by Weber-Wullf's typology as Pawn Sacrifice and Cut & Slide. |
| Syntax-preserving plagiarism | Makes use of synonym substitution and technical disguise, but still keeps the source's sentence structure and order. |
| Semantic-preserving plagiarism | Preserves only the meaning of the sentences and changes the words and structure, thus including not only paraphrasing but also translation. On that matter, they cite Velasquez et al. (2016) with whom they agree over translation being the ultimate form of paraphrase. |
| Idea-preserving or template plagiarism | Copies the structure of a source document or uses its concepts and ideas only. |
| Ghost-writing | The text itself is genuine, but written by a person other than the presented author of the document. Is thus undetectable through similarity-detection. |

Khan et al. (2018) distinguish between two broad categories of plagiarism that can be further detailed into common types of textual plagiarism that differ, regarding their definition, from other typologies describing similar types:

- **Literal plagiarism:** by introducing superficial or no alterations at all to the source text.
    - **Self-plagiarism:** the practice of recycling one's previous works in different research publications, without citation.
    - **Accidental plagiarism:** which can be perpetrated in any way, but in an unaware manner, or due to a lack of knowledge of the academic requirements.
    - **Clone plagiarism:** by replicating, as is, part of or the entirety of another's work, without citation.
    - **Mosaic plagiarism:** by keeping on the source sentences' grammatical structure, but applying superficial obfuscation techniques like synonym substitution and word omission.
- **Intelligent plagiarism:** which is more challenging to detect as it goes beyond simple text alteration, including:
    - **Structural plagiarism:** which goes a step beyond mosaic plagiarism, by also altering the grammatical structure.
    - **Metaphor plagiarism:** which occurs when a plagiarist takes a creative metaphor by an original author, and reuses the comparison while wording it differently.
    - **Idea plagiarism:** when the author does not steal text passages directly from a source but claims other's ideas that may be findings, conclusions, results, or original concepts.

Finally, Mozgoyov et al. (2010) discuss another typology that includes some types that we deem interesting in regard to plagiarism detection systems. It is summarized as such:

- Verbatim plagiarism.
- Obfuscation by paraphrasing.
- Intentional inaccurate use of references.
- Exploitation of the weaknesses of plagiarism detection systems.
- Tough plagiarism: which can be idea, structural or cross-language plagiarism. These are still hard to detect for both automated systems and humans.

### 2.2 Proposed typology

In this survey, we propose and consider a particular classification that is an aggregation of previously discussed typologies, but is more convenient for the study of similarity detection software. We distinguish between **content obfuscation plagiarism**, which modifies the text and challenges only the actual performance of the software, and **exploitation or technical obfuscation plagiarism** that makes use of technical alterations to bypass the detection system or exploit its breaches, regardless of its performance and ability to retrieve the sources. We further detail this typology in the following.

#### 2.2.1 Content obfuscation plagiarism
##### 2.2.1.1 Textual alteration plagiarism

This includes verbatim copying and all forms of alterations that can be made on the text content, while still keeping it semantically identical to its source. This includes paraphrase techniques like synonym substitution, re-ordering, change of tense and person, summarisation, and others. We decided to regroup all of these techniques into a single class, as it is unlikely for a plagiarist to use only one of them exclusively.

Complete paraphrase is still the hardest to detect, but with modern open technology, plagiarists can perform it with such ease that it becomes even more problematic. Paraphrase can be done automatically with specific online tools built for that purpose, which are available for free and gives convincing but varying results. Paraphrase can also be achieved through "Closed-Translation-Chains" or "Back- Translation": This technique makes use of machine translation services to translate a text from its original language to one or a series of languages, then back to its original language. With the current advancement in machine translation, the results given by such techniques are quite effective.

##### 2.2.1.2 Translation or Cross-language plagiarism

Not to be confused with back-translation, translation plagiarism is simply taking from a source that is in a different language than that of the submitted work, translating it by any means and then using it. While translation can arguably be considered a form of paraphrase, it still is far more complicated to detect since the gap between a source and a translated text cannot, at the moment, be traced back by any detection software, nor even search engines, as we will discuss in our results.

This is made even more complex since contribution to the research on the topic of cross-language similarity calculation is often restrained to the similarity assessment of two texts in different languages that are made already available, while the challenge in plagiarism detection in practice is also in retrieving that potential source text through information retrieval from search engines and different databases.

#### 2.2.2 Exploitation or visual and technical obfuscation plagiarism

Most of the exploits that are discussed in this paper are presented on the Internet as mere "tips" and ways to dupe plagiarism detection algorithms. This diminishes the ethical problems of such an act, as is the case in an article published on the website of a paper-mill service titled "Six Proven Ways to Cheat Turnitin" (Evans Connor, 2020).

While these underhanded techniques remain fairly simple to put into practice for the plagiarists, they can be astonishingly effective depending on the detection system being used for the task. Since a number of these tricks are just character-level manipulations, they can be performed in very little time with any text editor that offers the "Find and Replace All" functionality. The use of a number of these exploits, or "visual spoofing" means in plagiarism was investigated in further technical detail by Chow et al. (2016) in an attempt to apply classical cryptanalysis methods such as frequency analysis, the index of coincidence and the Chi-squared test to facilitate their identification as part of automated plagiarism detection.

In what follows are described the techniques that we consider in this paper, but we have no doubt that new ones will always be discovered, courtesy of the plagiarists' creativity:

   A. **Usage of homoglyphs**: which are characters from different alphabets that are visually similar but have different Unicodes, because of which they are not considered to be the same by software. Examples of this include using some characters from the Cyrillic alphabet to replace Latin characters, or from the additional characters of Pashto alphabet instead of standard Arabic ones, although the latter two are slightly visually different.
   B. **Insertion of micro-spaces**: this exploit is achieved by replacing single whitespaces between words with multiple whitespaces of a smaller font size, making the strings of text appear identical while being different in their code representation.
   C. **White ink**: an exploit in which the plagiarist inserts white-coloured, smaller letters between words. Those letters are totally invisible to the reader but are enough to make the whole text appear to the algorithm as a single string of gibberish that doesn't match any other text on any source.
   D. **Punctuation modification**: because some algorithms operate at sentence level, simply changing and moving the punctuation of the text may dupe them, giving different sentences than those of the source.
   E. **(In)voluntary misspellings or typos**: this is achieved by taking a correctly written text and altering some of its words into typos. This exploit is the most visible as it requires a large number of typos in a text in order to significantly lower the detection rate.
   F. **Text as Image**: where the plagiarist smartly replaces snippets of text in his work with a screenshot of that same text. The images are ignored during the analysis by most detection systems, as it requires the use of Optical Character Recognition (OCR) algorithms that would constitute an additional computational load for the detection task.

It is still worth noting that none of these methods are totally impregnable, and most of them can either be directly detected and countered or marked by the systems, or are made clear to the human reviewer through the output report, as can be the case with the white ink exploit if the text is displayed in a unified font colour in the report, or for text-as-image if the images are removed in the reports, leaving a large gap in-lieu of the image, which can grab the reviewer's attention.

### 2.3 Comparison of Plagiarism Detection Systems

Many comparative studies were conducted around plagiarism detection systems since the early 2000s, with varying scopes and criteria. The shared goal was to assess their suitability for academic use, either globally or in a specific context.

For the languages examined by those surveys, the focus is on English for most cases, or includes English besides one other language. Very few surveys consider multiple languages in a single comprehensive study. Research and surveys performed by Dr. Weber-Wullf[1] evolved towards that idea by considering German, Japanese, and English in the comparisons.

Even fewer surveys check the mark when it comes to assessing the performance of the systems at detecting similarity while providing detailed scores and, more often than not, they poorly include technical and functional criteria, which are necessary during comparison for a decision-making process by institutes willing to acquire access to such service.

Maurer et al. (2006) compared tools by considering functional criteria like the processing of tabular content and images, the ability to exclude certain sources, and the verification of the references. They also examined the detection performance against verbatim plagiarism, paraphrasing as well as translation plagiarism. They concluded that plagiarism detection systems fall short against paraphrasing and translation plagiarism, or when source content is not available on the Internet. They also perform poorly with tabular content and special characters. On the other hand, their methodology, as well as their testing corpora, were not detailed.

---

[1] All the surveys are listed under https://plagiat.htw-berlin.de/software-en

Ali et al. (2011) compared 5 tools in regard to their language support, their database scope (online and offline databases and the Internet) and the format of the sources (books, articles, magazines, PDF…), the ability to process multiple documents in parallel as well as the ability to check sentence structure and synonym. Their comparison yielded very close results for all systems and it is not explained on which basis the scores were given. Their methodology and their corpora were not reported either.

Vani and Gupta (2016) reported on 8 systems in regards to their main features, then they further investigated three among those tools as for their ability to detect 4 distinct levels and techniques of obfuscation: simple copy-paste, random obfuscation, back-translation (by machine-translating text from English to Hindi, then back to English) and finally by summarization. Their results have shown the detection to be accurate across systems when no obfuscation is applied, while it decreases significantly against other techniques, reaching a similarity score of 0% for all three tools when fed back-translated text.

Birkić et al. (2016) in their survey compared 4 tools by examining a number of technical and functional criteria besides the detection quality. Some criteria include the availability of an API (Application Programming Interface) or plugins for Learning Management Systems, the deployment options, the scope of source retrieval and the possibility to include personal content, the costs and licensing modals and finally the aptitude to be used in an organisational setting by discussing the available user roles and authentication modalities. This study is also interesting for its inclusion of some indicators which customises it for use in a particular academic context: Croatian universities and institutes. It considers the ability to authenticate with "AAI@EduHr e-ID", which is the authentication and authorisation standard in science and higher education in Croatia, and also the inclusion of the Croatian repositories DABAR and HRČAK. This study does not, however, consider the quality of detection in depth.

Nahas (2017) studied 15 tools, some of which are obsolete or not actively maintained. The study was limited only to a short description of each tool and recommendations for the users of those systems, and there was no actual comparison to distinguish the tools besides a classification into open (free) and commercial programs.

Shkodkina and Pakauskas (2017) provided an overview of the most common and serious forms of plagiarism around the world and in Ukraine in particular before performing a comparative study on 3 systems based on criteria that were identified as the most important for users in Ukrainian universities and by discussing the advantages and disadvantages of each one. Their criteria were divided into four subsets: affordability, material (format) support, functionality, and showcasing (the ability to establish the discussion between teachers and students). The methodology and criteria were ample and well documented but the scoring was binary (Yes or No) for almost all criteria, even when a score would have been more descriptive like for the paraphrase detection that was given a "Yes" for all three systems, which would drive to think that the systems are equally good at that task. Their corpora were also not presented.

Sobhagyawati (2017) and Jharotia (2018) each presented more than 30 and 10 systems respectively, for both textual and source code similarity detection. Their surveys did not aim to offer a comparison between said systems, but simply list, describe and report on the most used or referenced tools on the market, the free as well the commercial ones.

Chowdhury and Bhattacharyya (2018) compared 33 tools in an aggregation of multiple surveys previously conducted by authors. They assessed the tools in terms of user-friendliness and the ability to submit more than one file at once, the use of either an extrinsic (within a corpus) or intrinsic (based on the suspect document only) approach to detect plagiarism, and finally whether or not the tools are free, open-source, or commercial. They also considered some systems that are built around similarity detection in source code. Their survey is nevertheless an aggregation of other works that employ different methodologies and corpora to make their assessment, with a number of those surveys being out-dated and reporting on tools that may have evolved through the years, or became obsolete and unusable. Their list of compared tools also includes systems (Hawk-Eye, Maulik) that were not developed or made available to the public, but only published in research papers.

Members of the European Network for Academic Integrity (ENAI), including the earlier-mentioned Dr. Weber-Wullf, formed a group going by the name of "TeSToP" (for Testing of Support Tools for Plagiarism Detection) and published in 2020 what is arguably the most in-depth and elaborate comparative study of plagiarism detection software (Foltýnek et al., 2020). In their study, they considered eight distinct languages (English, German, Spanish, Italian, Turkish, Czech, Latvian, and Slovak) and compared no less than 15 web-based plagiarism detection systems. Their criteria aimed to provide a detailed examination of the quality of detection in each language, with consideration to different sources and types of online content and different obfuscation methods. Their usability evaluation was very exhaustive and included more than 20 criteria that users in an academic context can relate to.

In our study, we will attempt to complement their results by also considering French and Arabic languages, and examining in depth the efficiency of detection systems against obfuscation by technical exploits and translation plagiarism. We will also include several novel tools that were not included in any previous surveys

## 3. Methodology

To collect the data necessary to compare between plagiarism detection systems, comparison criteria were first selected, on the basis of which we will perform our evaluation. We then built a corpus of documents to use for the tests, enabling us to assess each system according to our criteria. After that, we listed and contacted all the systems' vendors individually, by requesting a free trial and explaining the goal and purpose of this survey.

All the tests were performed between November 2020 and March 2021. As these systems evolve and improve continually, the results that they yielded during this testing period may differ from the results that can be achieved at a later date, even by reusing the same dataset.

### 3.1 Comparison criteria

Our set of criteria addresses multiple aspects of the systems. We classified them into functional, technical, detection performance, robustness against plagiarism exploits, and finally usability and user experience. We provide a detailed list each subsection of criteria in what follows:

#### 3.1.1 Functional Criteria:

Includes the basic functionalities that are offered to the user, as well as more niche use cases that are relevant to consider in an academic context:

a) Supported languages,
b) Supported file formats,
c) Ability to batch-upload and process files,
d) Support for collusion detection: comparing multiple files against each other in a closed corpus. This is usually performed on students' assignments to detect cheating,
e) Ability to include the user's database of documents during comparison,
f) Ability to exclude and select sources against which similarity is checked: in order to exclude cases that are ascertained not to be plagiarism, like a valid citation, which would drastically but unfairly raise the similarity score,
g) Ability to let students submit their work directly into the platform, without the teacher needing to collect them through other means and submitting them manually.
h) Support for Optical Character Recognition and the ability to analyse files that were submitted as images, as well as images that are inserted within text files.
i) Support for tabular content in text files.

For the exception of the first three criteria (supported languages, file formats and file size), the rest will be scored as being either available or not.

#### 3.1.2 Technical Criteria:

This subset aims to assess how the systems perform in terms of portability, maintainability, and reliability. These are important factors as detailed by the ISO/IEC 9126 standard (International Organization for Standardization [ISO], 2001) that describes a software quality model composed of 6 main factors, which are to always take into account when acquiring or developing any type of software. For the case of similarity detection systems, we will try to evaluate that by examining:

1. The software's type and delivery model: Software as a Service (SaaS), On-Premise, or simple desktop software,
2. The integration with Learning Management Systems (LMS),
3. The availability of an Application Programming Interface (API) for integration in custom systems,
4. The average processing time,
5. The search engine used for candidate retrieval: as search engines are not equal in terms of websites and sources that they crawl and include in their results,

### 3.1.3 Detection performance

For the performance in detecting similarity cases, we assessed the plagiarism detection systems independently for each of the three languages considered in our study: English, Arabic, and French. We also consider three content sources: Wikipedia[2], other open-access web content like articles and blog posts, as well as open-access research papers. We excluded closed-access academic depositories from this study, but are an added value that will note in case they are handled by the system). Thus, for each pair of language and content source, we examine the following:

- The system's performance at detecting verbatim plagiarism,
- The system's performance at detecting paraphrase and synonymizing plagiarism,
- The system's performance at detecting translation (cross-language) plagiarism.

These criteria will be evaluated based on the average percentage of plagiarism reported by the systems, and will be given a score ranging from 0 to 10 accordingly.

### 3.1.4 Robustness against technical plagiarism exploits:

This subset of criteria aims to point out each system's exploits, as well as its robustness against technical plagiarism obfuscation techniques that may be used by plagiarists.

A system should either be able to detect and address these technical disguises automatically or at least help the reviewer in noticing them:

- Replacement of characters by homoglyphs in both Latin and Arabic languages.
- Insertion of micro-spaces between words.
- Usage of the white-ink technique.
- Modification of the source text's punctuation.
- Insertion of (in)voluntary misspellings or typos.
- Replacement of text by image-as-text.

Each criterion among this subset was scored as follows:

- 2 points are given to the system if it detects and bypasses the issue, or if it just notifies the user about it, or if it is not affected at all by it.
- 1 point is given if the system only makes the alteration visible to the reviewer, while not actually detecting it, and a drop in the similarity score can still be observed.
- 0 point is given otherwise.

The score for the (in)voluntary misspellings is ranges between 0 and 2 and represents how impacted is the calculated similarity score only, with 0 meaning that it is drastically decreased.

### 3.1.5 Usability criteria:

Functionalities offered by software and its quality of detection are not the only relevant factors that should be taken into account. It is important to consider that a large portion of users interested in plagiarism detection systems are teachers who may not be accustomed to the use of complicated or poorly intuitive interfaces, or may find them hard and inefficient to use. We assess the usability of the software in the following manner:

- A score out of 5 for general ergonomics and ease of use:
    - 1 point for the intuitiveness of the user interface,
    - 1.5 point for the navigation and the user experience,
    - 1.5 points for the workflow to upload a file and to access the reports
    - 1 point for the ability to upload files in batch,
- A score out of 5 for each of the online and downloadable reports that we examine individually:
    - 2 points for the intuitiveness of the report, its usage (for online reports only) as well as how results are presented,
    - 2 points for how similarity instances and their sources are showcased,
    - 1 point for the ease of navigation through the report and how much of the original submitted file's structure is preserved.

---

1. [2] https://www.wikipedia.org

It is worth noting that most usability criteria cannot be asserted with total objectivity, as each user may express a different opinion regarding an interface or his experience using it.

### 3.2 Corpus of documents used for the tests

To perform our tests in a manner that would guarantee a fair basis for comparison, we used the same corpus of documents across all the plagiarism detection systems involved in our study.

We built our corpus into two distinct subsets: The first is composed of standard text documents, made to assess the plagiarism detection performance, while the second is a set of singular documents used to determine the systems' reliability against the obfuscation techniques and exploits that we described in the previous section, as well as the availability and efficiency of some other functionalities.

### 3.2.1   Standard text documents

Documents in this set are instances of plagiarism cases that involve no obfuscation or textual obfuscation only. We divided three equal subsets for each of the three languages (English, Arabic, and French), with different sources each time, and took the average of three scores, allowing us to obtain less biased results.

For the origin of the text content, we used real content that was chosen at random and from three types of sources:

1. Wikipedia pages,
2. Web articles from online publishing platforms, news and scientific websites,
3. Open access research papers and thesis.

A single document in our corpus includes content from only one of these three types of sources, but in multiple instances. For example, we took samples from three different research papers and thesis in English to build one verbatim English research paper test document.

We considered Wikipedia separately from the other forms of web content and articles because some vendors[3] report scraping and including part of or the entire content of Wikipedia on their databases. This allows them to perform particularly well against that source and can be justified with the fact that Wikipedia remains one of the most recurring online sources of information among students (Lim, 2009).

As for research papers, we limited our scope to sources that are available in open access. This was done in order not to bind our tests to a particular closed-access repository that may be included in the database scope of one PDS but not the others, since they can be subjected to exclusive access, as stated by one of the vendors contacted for the survey performed by Foltýnek et al. (2020). Among the set of systems examined in our study, three of them use only open access content that is reachable through search engines.

---
[3] Docoloc stated to us that they include "all articles from Wikipedia in English German French and Spanish" in their internal database.

It is also to note that we excluded research papers in Arabic from our survey. This was decided after we found out that they are seldom crawled and indexed by search engines, and they are hardly searchable. Thus, we can easily take a whole paragraph from a thesis in Arabic and use it to query a major search engine, and it will still fail at retrieving it.

Each verbatim document was then altered in two ways: paraphrased using online services, and translated towards the two other languages, as detailed in table 6. We decided to use automated paraphrase and translation services because they became easy to use and widely accessible to the public. For the case of machine paraphrase tools, the outputted text is not always perfect in terms of understandability and fluency, and minor manual interventions were sometimes required, but the tools are still of value since they allow the paraphrasing of large volumes of text in mere seconds.

**Table 6: Levels of obfuscation for each source**

| Level of obfuscation | Method |
|---|---|
| Verbatim documents | text is taken as is from the sources, with no modification. |
| Automated paraphrase documents | the verbatim documents which we paraphrased automatically using online services like the Smodin[4] suite of tools |
| Automated translation documents | the verbatim documents from the other two languages, which we translated automatically using online services like Google Translate[5]. |

The composition of the corpora for standard text documents is illustrated in Figure 2.

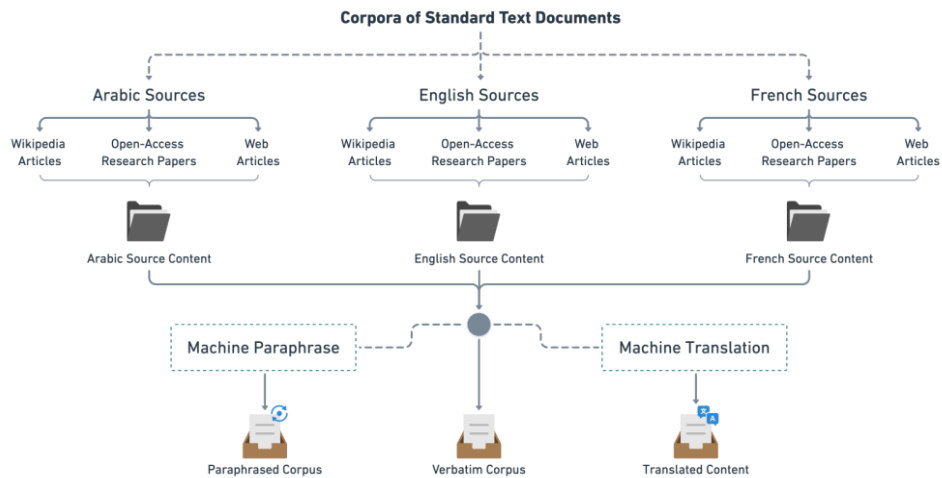

**Figure 2: Building methodology for the corpora of standard tests**

Files used for the tests were in two formats: PDF and DOC. We used the PDF format by default for documents in English and French languages. As for Arabic, we only used the DOC format since we noticed that almost all systems have trouble processing PDF files in Arabic. Upon further investigation, it appears to be an issue in the format itself and the encoding of Arabic text.

For the systems that did not offer file upload options, text was simply copied and pasted from the DOC files.

### 3.2.2 Particular test documents

With this subset, we investigate whether the systems are vulnerable to plagiarism techniques that do not modify the content text itself but introduce some form of noise that can disturb the systems' analysis and, eventually, drastically decrease the overall reported plagiarism score. Detailed explanations for all of these exploits are provided in the typology we presented in the previous sections. We prepared one file for each exploit and, in table 7; we explain how we built each file and provide an example of the modifications, most of which are indistinguishable.

---

[4] https://smodin.me
[5] https://translate.google.com

**Table 7: Technical disguise techniques and how they were performed**

| Technical exploit | Method |
|---|---|
| Homoglyphs for Latin languages | we substitute all instances of the Latin characters {a - o} by their Cyrillic similar counterparts {а - о} using the "Find and Replace" all functionality of the text editor.<br>**Example:** "In deep learning, each level learns to transform its input data into…" to "In deep learning, each level learns to transform its input data into…" |
| Homoglyphs for the Arabic language | we substitute all instances of the Arabic characters {ن - ر - ف - و} by their Pashto similar counterparts {ڼ - ړ - ڤ - ۇ} using the "Find and Replace" all functionality of the text editor.<br>**Example:** "...على الرغم من أن معظم الشعر في تلك الحقبة لم يتم الحفاظ عليه، فإن ما تبقى" to "...ﻋﻠﯽ ﺍﻟﺮﻏﻢ ﻣﻦ ﺃﻥ ﻣﻌﻈﻢ ﺍﻟﺸﻌﺮ ﻓﻲ ﺗﻠﻚ ﺍﻟﺤﻘﺒﺔ ﻟﻢ ﻳﺘﻢ ﺍﻟﺤﻔﺎﻅ ﻋﻠﻴﻪ، ﻓﺈﻥ ﻣﺎ ﺗﺒﻘﻰ" |
| White ink | we replace all spaces between words with a random letter then change its font colour to white, and reduce its font size in order to maintain visual similarity.<br>**Example:** "In deep learning, each level learns to transform its input data into…" to "In deep learning, each level learns to transform its input data into…" |
| Micro-spaces | we simply replace each instance of white space character by two instances, and reduce their font size in order to maintain visual similarity.<br>**Example:** "In deep learning, each level learns to transform its input data into…" to "In deep learning,  each level  learns  to transform  its input  data into…" |
| Modification of punctuation | we manually introduce additional punctuation characters and change the existing ones.<br>**Example:** "In an image recognition application, the raw input may be a matrix of pixels; the first representational…" to "In an image recognition application the raw input may be a matrix of pixels: The first representational…" |
| Intentional typos | we manually alter a random number of words by intentionally misspelling them.<br>**Example:** "In deep learning, each level learns to transform its input data into…" to "In deep laerning, each level learn to transform it's input data into…" |

This subset also includes special documents used to test other functionalities of the systems, as detailed in table 8.

**Table 8: Special test documents and how they were built**

| Criteria | Method |
|---|---|
| OCR and figures handling | by taking a screen capture of a text page and putting the picture in a text file. |
| Processing of tabular data | by copying tables from 4 different sources (2 from Wikipedia, 2 from research papers) and examining how they are parsed by the systems. |
| Collusion detection | by producing 3 files that are all a different mix and paraphrase of the same source, simulating the distinct homework of 3 cheating students. We then examine how collusion is handled and presented by the systems. |
| Quotes and citations | by using quotes and different styles of citations to see how the systems handle them and whether they are excluded from verification. |

### 3.3 Plagiarism Detection Systems involved in this survey

This section serves to present the plagiarism detection systems that were included in the testing phase. Our primary selection criterion was for the tools to support at least one of the three languages considered in our study (English, French, and Arabic). 34 systems were on our initial list of candidates, this number was eventually reduced to 8 tools as the others were excluded for one or multiple reasons that include: the company not offering free trials, not responding to our request or emails, the similarity detection system or feature being down during our testing period, and there was even one company that apparently ceased to exist after our first attempt to reach out to them.

The information we provide in the following is based on what is mentioned on each vendor's websites, or what was quoted to us by their representatives during email exchanges or calls.

### Compilatio[6]

Is operated by the French company of the same name since 2005, who claims to be an undisputed leader in French-speaking markets of plagiarism detection, as well as the most used system among universities in France. Compilatio describes itself as "*An online anti-plagiarism software that detects similarities with thousands of digital documents and provides clear analysis reports*". It compares against a database comprised of open-access content, the customer's own previously submitted documents and documents submitted by all partner institutions using Compilatio Magister, which is their license for organisational use.

### Copyleaks[7]

Copyleaks describes itself as a plagiarism checker for both academia and businesses, is powered by artificial intelligence and can "*Detect plagiarism, paraphrased content, and similar text using sophisticated Artificial Intelligence (AI) based algorithms in 100+ languages*". The system is operated by Copyleaks Technologies, LTD, a company founded in 2015 based in the United States. Besides textual plagiarism detection, it offers an automated grading tool, the ability to compare entire websites against each other, as well as plagiarism detection in source code.

### Docoloc[8]

Presents itself as a plagiarism search tool that can be used to find similarities between text documents on the Internet. Docoloc is maintained by the German company Docoloc UG & Co since 2006. The tool is limited for usage by institutional customers only, as it cannot be used by private individuals, or to check single documents. Docoloc reported having a comparison database that consists of more than 25 million documents from open repositories like arXiv doaj.org, core.ac.uk, PMC, Hindawi and Dovepress, as well as all articles from Wikipedia pages in English, German, French, and Spanish. It also includes more than 10 million other freely accessible documents that were found as sources during the source retrieval phases of previous plagiarism checks for their users.

### DupliChecker[9]

A freemium plagiarism checker that is part of a suite of tools, including a grammar checker and a paraphrasing tool. The company behind it appears to be located in the United Kingdom and was founded in 2007, with the statement that they provide "*quality web content, proofreading and editing of content already placed on client's website*". The free tier of the plagiarism checker allows a maximum of 1000 words per unique query, and up to 10,000 words with a paying pro license. It checks only against content that is openly accessible on the Internet.

### Google Originality Reports[10]

Originality Reports is not a standalone similarity checker, but a feature that was introduced to Google Classroom and Assignments in 2019. It declares to use "*the power of Google Search to help students properly integrate external inspiration into their writing – while making it easy for instructors to check for potential plagiarism*". Originality Reports allows students to check their works against plagiarism up to 3 times before submitting a final version of their documents to their reviewing teacher.

This feature can be enabled free of charge in up to 5 assignments per class for organisations having the "Google Workspace for Education Fundamentals" license, with no limit on the number of students that can submit their files, but with a limit of 20 files per student (as of the period of tests). A paying upgrade is available that offers unlimited use and the ability to compare against documents in a domain-owned repository of past submissions.

---

[6] https://www.compilatio.net
[7] https://copyleaks.com
[8] https://www.docoloc.de/plagiat_anleitung.hhtml
[9] https://www.duplichecker.com
[10] https://edu.google.com/products/originality/

### Oxsico[11]

Short for Oxford Similarity Checker for Organisations, is a relatively new similarity checker managed by the Lithuanian company "Lingua intellegens", which has previous presence in the field of plagiarism detection through the "Plagramme" plagiarism checker. Oxsico claims to have the widest support for languages, with 129 different languages that can be processed by the software, and declares having the quickest time of comparison among competitors with an average of 7 seconds per page. Its comparison database includes more than 80 million scholarly documents. The system also promotes the use of contract cheating technology that is based on stylometry in order to determine authorship, which is available as a separate operation on the user's dashboard.

### PlagiarismCheck.org[12]

Is operated by the British company *Teaching Writing Online* that serves more than 77,000 users worldwide since 2011. The system is presented as a plagiarism-checking tool that helps to detect similarities and produces unbiased results. It promotes Its feature of *"detecting not only rearrangements in word order, but also in flagging substituted words with synonyms"* along with detecting changes in sentence structure and hidden symbols, and can also recognize correctly formatted quotes, which allows it to reduce false positives.

### StrikePlagiarism[13]

Describes itself as an "*academic anti-plagiarism system*". StrikePlagiarism is operated by the Plagiat.pl company, located in Poland. They highlight cooperating with more than 700 universities, and supporting more than 200 languages. The system is said to use an algorithm that can capture paraphrasing in any language, as well as several technical text manipulations. The database of documents used during comparisons is stated to include scientific journals and articles indexed in Scopus, Springer and Web of Science (among others), more than 3 million research papers from open access repositories like arXiv, Paperity and Termedia, as well as all documents submitted by institutions using StrikePlagiarism and agreeing to be part of their inter-university database exchange program.

We report in table 9, for each system, the supported languages as quoted on the vendors' websites, their licensing model, as well as the pricing criterion that is used.

Table 9: Language support, pricing and licensing comparison

| System | Supported languages | Pricing | Licensing |
|---|---|---|---|
| Compilatio | "More than 40 languages (including all Latin languages)" | per student | Proprietary |
| Copyleaks | "All unicode languages including Asian character languages" | per page | Proprietary |
| Docoloc | Not specified | per page | Proprietary |
| DupliChecker | "Various languages, including Spanish, Russian, Portuguese, Dutch, Indonesian, Italian, and Arabic" | per word /operation[1] | Freemium |
| Google OR | Not specified | unlimited | Freemium (for academic use) |
| Oxsico | "129 languages including writing S. like Greek, Latin, Arabic, Aramaic, Cyrillic, Georgian, Armenian, Brahmic family scripts, Ge'ez script, Chinese characters and derivatives (including Japanese Korean Vietnamese), Hebrew" | per student | Proprietary |

---

[11] https://www.oxsico.com
[12] https://plagiarismcheck.org
[13] https://strikeplagiarism.com

| | | | |
|---|---|---|---|
| PlagiarismCheck.org | "Works best with the English language, other languages are also supported" | per page | Proprietary |
| StrikePlagiarism | "German Spanish Portuguese Italian French Norwegian Swedish Dutch Russian Arabic Hebrew Greek Turkish Vietnamese Philippine Indonesian Hindu Ukrainian Bulgarian Romanian and many more" | per file | Proprietary |

[1] For the case of Duplichecker, when a user buys a license, it is limited not only by the number of words that will be processed in total, but also by the number of operations that it offers, with each operation having a word threshold as well. For this survey, we tested the free version of DupliChecker, and had to manually dividing each of our files into portions or sets of paragraphs having less than 1000 words and then calculating the global similarity score for the file.

The metric of "Page" is used to describe tokens that can be purchased, with each token equalling either 225 or 250 words depending on the system. A single page in a submitted file can thus be counted for more than one page unit when processed by a plagiarism detection system because the number of words per page in any file depends on a number of factors, including the document's layout and the font size.

## 4. Results and discussion

### 4.1 Functional criteria

The functional coverage of each system was assessed with a set of criteria that are summarised in table 10.

Most systems allow users to include their documents as possible sources for plagiarism detection. These can either be documents that were previously submitted for analysis, or documents that were purposefully uploaded to constitute an institutional repository.

As for collusion detection, Only Copyleaks offers the possibility to compare a set of documents in a closed corpus. For instance, this can be used to analyse the essays of a whole class of students against each other and output a single comprehensive report that would make it easy for the teacher to detect cheating attempts.

It is still worth noting that this can be achieved to an extent by all systems that offer the functionality of a user database, as sources can include other previously submitted files, but the teacher has to process them individually to consolidate all the results and identify clusters of students that cheated similarly, which can be hardly feasible when the number of individual files gets larger.

Table 10: Evaluation of the functionl criteria

| | User's database | Collusion Detection | Sources exclusion | Student submission | OCR | Tabular data |
|---|---|---|---|---|---|---|
| Compilatio | ✓ | ✗ | ✓ | ✓ | ✗ | 25% |
| Copyleaks | ✓ | ✓ | ✓ | ✗ | ✓ | 75% |
| Docoloc | ✓ | ✗ | ✗ | ✗ | ✗ | 25% |
| DupliChecker | ✗ | ✗ | ✗ | ✗ | ✗ | 0% |
| Google OR | ✓[1] | ✗ | ✗ | ✓ | ✗ | 25% |
| Oxsico | ✓ | ✗ | ✓ | ✓ | ✓ | 25% |
| PlagiarismCheck.org | ✓[2] | ✗ | ✓ | ✓ | ✗ | 0% |
| StrikePlagiarism | ✓ | ✗ | ✓ | ✗ | ✗ | 25% |

[1] Requires the upgraded license "Teaching and Learning Upgrade and Education Plus", and is not available in the free license "Google Workspace for Education Fundamentals".

[2] Is done on request, with an additional fee of 0.16$ per page added to the database.

Allowing the students to submit their files directly on the platform greatly helps the teacher, by eliminating the need to collect files through other mediums before submitting them. On the systems we tested, this functionality was presented in three different forms:

1. The students are also users of the system, and the teacher creates a submission form for specific students or group of students (Google OR, Oxsico, PlagiarismCheck.org).

2. The students do not need to be users of the system, and the teacher creates a submission form on the platform with a shareable link (Compilatio, Oxsico).
3. The teacher only shares an email address, which is generated at random by the system, and the students simply send their files by email to that address (Oxsico).

The ability to exclude sources from the similarity calculation is crucial, as it allows the reviewer to exclude irrelevant cases that can be false positives. This can have a drastic impact on the overall similarity score. More than half the systems offer that functionality. Google OR only allows excluding all citations at once but does not give control over sources.

The possibility to perform Optical Character Recognition (OCR) and process scanned image files containing text is supported only by Copyleaks and Oxsico. This functionality can find its use when in need to add to the checking database an old file, that was only digitized, in order to compare against it. It can also be a means to circumvent an obfuscation attempt by converting whole files into image format.

As for the processing of tabular data, all systems were capable of detecting, at most, only one table from the four we submitted, except for Copyleaks that processed and retrieved the sources for three out of them. It is still worth noting that Compilatio and Oxsico, just like Copyleaks, offer support for the Excel Spreadsheets format (XLS), but they do not process the tabular data in standard text files any better.

Finally, for the supported file formats, it is shown in table 11 that all systems, except one, can process the major text file formats (PDF, DOC and TXT).

**Table 11: Supported file formats for each system**

|  | DOC | PDF | TXT | ODT | RTF | LaTeX | Others |
|---|---|---|---|---|---|---|---|
| Compilatio | ✓ | ✓ | ✓ | ✓ | ✓ | ✓ | PPT, XLS, ZIP |
| Copyleaks | ✓ | ✓ | ✓ | ✓ | ✓ | ✓ | PPT, XLS, ZIP, HTML, Pages, Image formats, CSV, EPUB |
| Docoloc | ✓ | ✓ | ✓ | ✓ | ✓ | ✗ |  |
| DupliChecker | ✓ | ✓ | ✓ | ✓ | ✓ | ✓ |  |
| Google OR | ✗ | ✗ | ✗ | ✗ | ✗ | ✗ | Google Docs |
| Oxsico | ✓ | ✓ | ✓ | ✓ | ✓ | ✓ | PPT, XLS, ZIP, HTML, Pages, Image formats, Google Docs, |
| PlagiarismCheck.org | ✓ | ✓ | ✓ | ✓ | ✓ | ✗ |  |
| StrikePlagiarism | ✓ | ✓ | ✗ | ✗ | ✗ | ✗ |  |

Google Originality Reports detects plagiarism only in files in the Google Docs format. This is cumbersome as it forces the user to first upload his document into Google Drive, then open it with the Google Docs word processor to convert it into Docs format, before finally submitting it in Google Classroom or Assignment.

Most systems support other formats that are less commonly used like ODT and RTF. The LaTeX format, which is amply used in academic and scientific publications, is supported by half the systems. Compilatio, Copyleaks, and Oxsico are the only ones to accept compressed ZIP files, which can avoid users the need to upload large files and leverages Compilatio's smaller supported file size of 20Mb. At last, Copyleaks and Oxsico are the systems that support the most file formats

## 4.2 Technical Criteria

All the systems considered in our survey are distributed as SaaS (Software as a service), they are all web-based applications that only require the user to have a browser, which relieves the institutions from deployment and maintenance concerns.

With the exception of DupliChecker and Google COR, all systems offer APIs that can be used by universities to develop custom systems or LMS that incorporate plagiarism detection. The performance and quality of detection are reported to be identical, independently of the use of an API or the web interface. The difference lies in how easy it is to access the APIs, as some systems like Oxsico offer the possibility to generate and manage API keys directly from their dashboards, while others like Compilatio offer them as an additional service that is priced separately.

Similarly, integrations into Learning Management Systems are offered by all systems except Docoloc, DupliChecker, and Google OR. The Moodle platform appears to be the one that is most supported, followed by Canvas, Blackboard, and Google Classroom. This is summarized in table 12.

**Table 12: Evaluation of the technical criteria**

|  | API availability | LMS integration |
| --- | --- | --- |
| Compilatio | ✓ | ✓ |
| Copyleaks | ✓ | ✓ |
| Docoloc | ✓ | ✗ |
| DupliChecker | ✗ | ✗ |
| Google OR | ✗ | ✗* |
| Oxsico | ✓ | ✓ |
| PlagiarismCheck.org | ✓ | ✓ |
| StrikePlagiarism | ✓ | ✓ |

*For the particular case of Google OR, it is worth restating that it is by default a plugin integrated into Google Classroom and Google Assignments, and cannot be used independently, but Google Assignments can indeed be integrated into other LMS.

Another technical criterion we attempted to include in this comparison was the search engines that were used by each system during their source retrieval phases, as we suspected that using different engines would correlate with the quality of detection since the major search engine (Google) vastly exceeds the other ones in terms of number of indexed web-pages[13]. We eventually decided not to report on that, the reason being that not all the vendors were open about this information, stating that it was a corporate secret. It would thus be unfair to share about the technologies used by only some systems while others are guarding theirs.

Finally, as we were attempting to measure the average time taken by each system to perform the analysis, we eventually concluded that we cannot assess this consistently. This was due to the fact that the time taken by each system to process the same file can differ. It is expected for the systems to handle their users' queries in a queue, or parallel queues, which may extend the processing duration in times when there is a large number of simultaneous queries. Systems that have access to large computational resources are able to reduce this issue.

Compilatio does, in fact, explicitly report to the user that his analysis may take longer than usual during busy periods, by precisely indicating how much time would be spent in a waiting queue, and how much time would be required for the actual analysis.

It can then be stated that processing time can differ greatly across systems, but it also depends on other factors that are not in the user's control. In normal circumstances, the file size is the factor that mostly influences analysis duration.

### 4.3 Detection performance

#### 4.3.1 Verbatim plagiarism

Table 13 presents the average scores across all the participating systems for the verbatim plagiarism detection, given for each language and source pair.

It can clearly be noticed that all systems detect Wikipedia sources particularly well and regardless of the language. This supports our decision of considering Wikipedia separately from other web content as it would have biased our results. On the other hand, open access research papers and thesis are less well detected, this may indicate that plagiarism from sources that are files, and not text that is directly on web pages, is more challenging to detect since it adds an extra layer of parsing and processing.

As for the languages, English is the overall most supported one, and, defying the general expectations, Arabic is clearly not as under-supported as it is thought to be, with some systems showing similar performance for both French and Arabic.

We can make the assumption that the bottleneck when detecting verbatim plagiarism is the candidate retrieval phase of the system's algorithms, since all scores inferior to *7.0* during our tests correlated with at least one of our three sources not to be retrieved at all. This may also be tied to the search engines in use by the systems. Systems depending on engines with less populated indexes may have a disadvantage in that area.

Finally, it is observable that Copyleaks, Google OR, and Oxsico perform remarkably well across all three languages and types of sources.

**Table 13: Evaluation of the detection of verbatim plagiarism**

|  | English | | | French | | | Arabic | |
| --- | --- | --- | --- | --- | --- | --- | --- | --- |
|  | Wiki | W.A. | R.P. | Wiki | W.A. | R.P. | Wiki | W.A. |
| Compilatio | 9.0 | 10 | 4.0 | 10 | 6.0 | 4.5 | 10 | 7.0 |
| Copyleaks | 10 | 10 | 7.5 | 10 | 10 | 9.5 | 10 | 9.5 |
| Docoloc | 9.0 | 8.5 | 8.0 | 10 | 8.0 | 6.5 | 9.0 | 4.5 |
| DupliChecker | 9.0 | 7.5 | 5.0 | 7.5 | 5.5 | 4.5 | 7.0 | 4.0 |
| Google OR | 10 | 9.0 | 8.0 | 10 | 10 | 8.5 | 8.5 | 8.5 |
| Oxsico | 9.5 | 10 | 9.0 | 10 | 8.5 | 9.0 | 10 | 9.0 |
| PlagiarismCheck.org | 10 | 10 | 6.0 | 10 | 8.0 | 3.0 | 10 | 7.5 |
| StrikePlagiarism | 10 | 10 | 5.5 | 10 | 8.0 | 5.0 | 5.5 | 5.5 |

- Wiki: sources are Wikipedia pages
- W.A: Web Articles, sources are web articles on news and scientific websites and publishing platforms.
- R.P: Research Papers, sources are open access research papers and journal articles.

#### 4.3.2 Paraphrase plagiarism

In table 14 are shown the average scores of detection for source text that was paraphrased using online services. It is worth restating that the original content we obfuscated for this purpose is the exact same content used to assess the systems' performance against verbatim plagiarism.

It is shown that systems perform poorly when the text undergoes syntactic and lexical alterations, regardless of the languages in use. Copyleaks is the only exception as it was able to detect and highlight, to an extent, paraphrased content from Wikipedia articles in both French and English.

For the sources, web content that is directly searchable on web pages like Wikipedia and web articles, have a very slight advantage as their sources were sometimes correctly retraced by some systems, but poorly matched.

This can be explained by the fact that paraphrased text can still include short passages that are unchanged, through which the systems can retrace the source, but still fail at matching it with the paraphrased text. We can thus assume that both the similarity calculation and candidate retrieval phases of the plagiarism detection systems are challenged when facing obfuscation by paraphrase.

**Table 14: Evaluation of the detection of paraphrase plagiarism**

|  | English | | | French | | | Arabic | |
| --- | --- | --- | --- | --- | --- | --- | --- | --- |
|  | Wiki | W.A. | R.P. | Wiki | W.A. | R.P. | Wiki | W.A. |
| Compilatio | 0 | 0 | 0 | 1.0 | 1.5 | 0.5 | 0 | 0 |
| Copyleaks | 6.0 | 4.0 | 0.5 | 4.5 | 2.5 | 2.5 | 0.5 | 1.0 |
| Docoloc | 1.5 | 1.0 | 0.5 | 2.0 | 2.0 | 1.0 | 0 | 0 |
| DupliChecker | 0.5 | 0.5 | 0 | 0.5 | 1.5 | 0.5 | 0 | 0 |
| Google OR | 3.0 | 2.5 | 1.5 | 1.5 | 2.0 | 0 | 2.0 | 2.5 |
| Oxsico | 2.5 | 1.5 | 2.0 | 1.5 | 1.5 | 1.0 | 0 | 0.5 |
| PlagiarismCheck.org | 2.0 | 0.5 | 0.5 | 2.5 | 3.0 | 0 | 1.0 | 0 |
| StrikePlagiarism | 2.5 | 3.5 | 1.0 | 1.0 | 1.0 | 0 | 0.5 | 0 |

### 4.3.3 Translation plagiarism

Tables 15, 16, and 17 aggregate the similarity detection results of the systems against translation plagiarism across all pairs of languages:

- From French and English to Arabic,
- From Arabic and English to French,
- From French and Arabic to English.

The source content used for these tests is still the same that was used for the verbatim plagiarism detection, it was simply translated to the target languages.

It can be observed that cross-language plagiarism detection is not supported at all by any of the systems, as none of them could retrieve a source that was in the original, non-translated language.

In some cases, systems reported sources that match the submitted text, but these are always just instances of translated text that is available on other web pages as we will detail for each of the three tables.

Table 15 shows the results of submitting documents in Arabic that were translated from English or French sources. The only slightly noticeable similarity scores are from English Wikipedia content. We then found out that the content of one Arabic Wikipedia page that we used was a mere translation of its English version. The retrieved source was still the Arabic page, and not the English page.

In table 16 are presented the results of submitting documents in French that were translated from English or Arabic sources. The detection rate for content translated from English Wikipedia pages was surprisingly high. By examining the sources, we identified a mirror version of Wikipedia linked as "Qaz.wiki". All the content on this mirror version was built by automatically translating the English Wikipedia pages into other languages, including French. We also uncovered that, within our French content, two web articles were translations of the same articles on their websites' English localised versions, which explains the slight detection scores on the column for web articles from English sources. Those articles were respectively taken from a news and a scientific website.

Finally, Table 17 shows the results of submitting documents in English that were translated from French or Arabic sources. No particular results can be noted, except that some web articles in Arabic within our collected corpus were plagiarised from similar English articles available on the web. Translating those articles back to English allowed for similarities with the original content to be detectable.

In the end, we conclude that all plagiarism detection systems covered in this survey operate on the submitted document's language only, and any sources detected against cross-language plagiarism are just instances of translation that already exist on the Internet.

Table 15: Evaluation of the detection of translation plagiarism in Arabic text

|  | from English sources | | | from French sources | | |
|---|---|---|---|---|---|---|
|  | Wiki | W.A. | R.P. | Wiki | W.A. | R.P. |
| Compilatio | 0.5 | 0 | 0 | 0 | 0 | 0 |
| Copyleaks | 0.5 | 1.0 | 1.0 | 1.0 | 0.5 | 0 |
| Docoloc | 1.0 | 0 | 0 | 0 | 0.5 | 0 |
| DupliChecker | 0.5 | 0.5 | 0 | 0 | 0 | 0 |
| Google OR | 2.5 | 0.5 | 0.5 | 0.5 | 0 | 0 |
| Oxsico | 1.5 | 1.0 | 0 | 0 | 0 | 0 |
| PlagiarismCheck.org | 1.5 | 1.5 | 0 | 0 | 0 | 0 |
| StrikePlagiarism | 0.5 | 0 | 0 | 0 | 0 | 0 |

Table 16: Evaluation of the detection of translation plagiarism in French text

|  | from English sources | | | from Arabic sources | |
|---|---|---|---|---|---|
|  | Wiki | W.A. | R.P. | Wiki | W.A. |
| Compilatio | 4.0 | 1.5 | 0.5 | 0 | 0.5 |
| Copyleaks | 9.0 | 5.0 | 1.0 | 1.0 | 0.5 |
| Docoloc | 5.0 | 1.0 | 1.0 | 1.0 | 1.0 |
| DupliChecker | 4.5 | 2.0 | 0.5 | 0.5 | 1.0 |
| Google OR | 9.0 | 3.5 | 1.0 | 0.5 | 0 |
| Oxsico | 6.5 | 2.0 | 1.0 | 1.0 | 1.0 |
| PlagiarismCheck.org | 5.5 | 1.5 | 0 | 0 | 0 |
| StrikePlagiarism | 6.0 | 3.0 | 0 | 0 | 0.5 |

Table 17: Evaluation of the detection of translation plagiarism in English text

|  | from French sources | | | from Arabic sources | |
|---|---|---|---|---|---|
|  | Wiki | W.A. | R.P. | Wiki | W.A. |
| Compilatio | 0.5 | 0.5 | 0.5 | 0 | 0.5 |
| Copyleaks | 2.0 | 2.0 | 1.5 | 1.0 | 1.5 |
| Docoloc | 1.0 | 1.5 | 1.0 | 0.5 | 1.5 |
| DupliChecker | 0 | 0.5 | 0.5 | 0.5 | 0.5 |
| Google OR | 1.0 | 1.5 | 1.0 | 1.5 | 2.0 |
| Oxsico | 1.0 | 1.5 | 1.5 | 1.5 | 1.5 |
| PlagiarismCheck.org | 1.0 | 1.0 | 0 | 1.5 | 1.0 |
| StrikePlagiarism | 0.5 | 1.0 | 0.5 | 0 | 1.5 |

### 4.4 Robustness against technical plagiarism exploits:

The results of using technical ways to hide plagiarism are reported in Table 18.

It shows that inserting micro-spaces between words and changing the punctuation does not affect the detection, except for Google OR, because its detection score dropped by half by simply altering the text's punctuation. This behaviour may indicate the use of a sentence tokenization algorithm that depends too much on punctuation marks.

It is worth adding that, despite showing a similarity score of only 54%, Google OR still highlighted all the passages. This may be considered as a caveat since it drives the user not to trust the global score to assess how much of the text was highlighted.

For the usage of homoglyphs in Latin languages (English and French), Oxsico was able to detect the unusual characters, change them back to their original forms and proceed to the analysis without issues. Copyleaks and StrikePlagiarism were able to identify and highlight those characters but not work around them, as the reported similarity score decreased for both of them. Compilatio did not seem to detect nor highlight homoglyphs, but it was not as much affected as the rest of the systems, which outputted a similarity score of 0%.

As for homoglyphs in Arabic text, the same behaviour was shown by StrikePlagiarism. Google OR was not able to identify the Pashto characters but they were made even more visible in the reports, and the similarity scores did not drop as significantly as for the other systems, which either reported a score of 0% or did not recognise the language at all, as pointed by Compilatio in an error message.

For the use of the technique of white ink, both PlagiarismCheck.org and StrikePlagiarism were able to completely ignore the invisible letters, with the latter even reporting them. The other systems were not able to disregard them, but the white letters were made visible in the reports because the font colour was unified to black, which can attract the reviewer's attention. Copyleaks and Oxsico were the only systems that did not detect nor made them visible since they preserved the format of submitted documents.

Finally, Copyleaks and Docoloc were the systems to be the least affected by the extensive and intentional introduction of typos, which may indicate their use of a fuzzy matching algorithm when comparing texts. This obfuscation technique is still the least effective, as the resulting text is full of typos in a way that may attract more negative attention from the reviewer than plagiarism.

Table 18: Evaluation of the robustness against exploits plagiarism

|  | Homoglyphs in Latin | Homoglyphs in Arabic | Micro-spaces | White ink | Punctuation | Typos |
|---|---|---|---|---|---|---|
| Compilatio | 1 | 0 | 2 | 1 | 2 | 0 |
| Copyleaks | 2 | 0 | 2 | 0 | 2 | 2 |
| Docoloc | 0 | 0 | 2 | 1 | 2 | 1.5 |
| DupliChecker | 0 | 0 | 2 | 1 | 2 | 0 |
| Google OR | 0 | 1 | 2 | 1 | 0 | 1 |
| Oxsico | 2 | 0 | 2 | 0 | 2 | 0.5 |
| PlagiarismCheck.org | 0 | 0 | 2 | 2 | 2 | 0.5 |
| StrikePlagiarism | 2 | 1 | 2 | 2 | 2 | 0 |

### 4.5 Usability criteria

Table 19 shows the consolidated usability results for the systems.

As users, we noticed that offline and downloadable reports are usually different from their online counterparts. This is because online reports are not files per se, but web pages that offer an interactive experience and more ease of use, whilst they are downloadable in PDF format only.

The main challenge of offline reports resides in how to visually link highlighted passages with retrieved sources while preserving the structure and layout of the submitted file and the ease of navigation through the report.

The only clear distinction is Docoloc, which solves this issue by offering only one format for both reports: an HTML file. By opting for this format, users can interact with reports even after downloading them. The interaction consists of:

- highlighting passages from the text that are associated with a source selected by the user from the list,
- showing, in a popup, all retrieved sources for a passage on which the user clicks.

It is still worth noting that Docoloc does not offer the ability to exclude sources.

Oxsico offers the best downloadable reports because it preserves the original file structure and only annotates the text by highlighting the detected cases, and only links them to the most prominent source, while still listing all the other sources.

As for the other systems' downloadable reports, they mostly lack readability because of the intensive use of annotation as well as the breakdown of original structure and layout.

As for the ergonomics, we note the following points for each one of the sub-criteria:

- **Intuitiveness:** Plagiarism detection systems are usually intuitive to use with no particular issues.
- **Navigation and User Experience:** The ease of navigation and the user experience are tied with the modernity of the user interface, as users have built habits and became accustomed to using modern and fluid software and applications. Docoloc and DupliChecker were the only systems with a distinguishable lack in this area.
- **Workflow:** for the workflow of uploading files and consulting reports, we faced particular issues only with Duplichecker and Google Originality Reports. With DupliChecker, we had to manually split texts into portions that were less than 1000 words, then calculate a global score for each file, and the reports were not saveable to a user account. As for Google OR, a teacher cannot upload files himself unless he uses a secondary account, as only students can submit files for a given homework. The reports are hard to find and it does not show how much time is required for the analysis. which may be problematic as we had to wait for multiple hours when submitting large files like a thesis. Finally, the bigger downside is the obligation to first convert each file into the Google Docs format before submitting it, as this is the only format supported by this system.
- **Batch Processing:** All systems offer the ability to batch-process files except for DuplichCheker. We also faced a lot of difficulty with using this functionality with StrikePlagiarism since it forces the user to download a template spreadsheet file and fill an entry for each file, by providing all the metadata and the file name, then submitting both the spreadsheet and a compressed .zip folder containing all the files. This was very cumbersome and we eventually found it to be much easier to simply upload files one by one.

**Table 19: Evaluation of the usability criteria**

|  | Ergonomics | Online reports | Offline reports |
|---|---|---|---|
| Compilatio | 4,25 | 2,5 | 2,75 |
| Copyleaks | 5 | 5 | 3.75 |
| Docoloc | 3,5 | 4,5 | 4,5 |
| DupliChecker | 2,25 | 1,5 | 1,25 |
| Google OR | 3,75 | 4 | 2 |
| Oxsico | 5 | 5 | 4.25 |
| PlagiarismCheck.org | 4,25 | 4 | 2,5 |
| StrikePlagiarism | 4,5 | 4 | 3,5 |

## 5. Recommendations

Through this survey and the collected results and information, we were able to detect several problems regarding plagiarism detection systems, some of which are more critical than the others. Based on that, we suggest the following as potential improvements points for said systems:

- Invest in more advanced plagiarism detection algorithms that can detect not only verbatim plagiarism but also paraphrase, by making use of the modern machine learning and natural language processing techniques.
- Attempt to handle the issue of cross-language plagiarism, especially from English towards other languages, as it is very easy nowadays to copy and translate from sources that are in a different language without being detected.
- Take into account the non-textual plagiarism exploits that are very easy to perform by plagiarists and can pass unnoticed and drastically change the outputted results.
- Include other functionalities that would help users in academic contexts, like collusion detection reports and allowing students to submit their files directly into the platform.
- Offer better support for the Arabic language, as well as the PDF format for files in Arabic.
- Handle citations and references more accurately in order to reduce not only false positives, but also false negatives by not considering every text between parentheses to be a correct reference.

It is worth noting that some systems, among the ones we tested, already implement a number of the aforementioned suggestions.

We also support the following recommendations that were suggested by Foltýnek, et al (2020)

- Dropping the single overall similarity score in favour of metrics that are more specific and less prone to be misleading during decision-making.
- Offer reports that are more intuitive and understandable, and that clearly showcase detected passages and their retrieved sources.
- Avoid burdening the user's workflow by not supporting batch upload or forcing the input of metadata.

## 6. Conclusion and future work

Through this survey, we overviewed several typologies of plagiarism previously proposed and discussed by authors before building a novel typology that we deemed more suitable for the comparison of plagiarism detection systems. We then examined 8 distinct systems based on well-defined criteria and a corpus built for the purpose. We focused on French and Arabic, besides English, to contribute to the study of these under-represented languages that are widely used in education and academia worldwide but are seldom considered or included in the comparisons of plagiarism detection systems. We also paid special attention to non-conventional plagiarism techniques that are very easy to apply by plagiarists but have the potential to drastically drop the detection rates as we uncovered on several systems. In general, we noted that plagiarism detection software greatly evolved during recent years in terms of functionalities, language coverage, usability, and source database scope, but we also found out that plagiarism by paraphrase is still under-developed, albeit some systems are already advancing into that path, while cross-language plagiarism remains an uncharted field and an easy opportunity for plagiarists to exploit.

This survey can be further elaborated by including other similarity detection systems and develop bigger corpora in order to reduce an eventual bias. Besides plagiarism techniques, it would also be extremely interesting to study the plagiarism detection techniques that are used by the systems to find out which ones are more useful in practice.

## Acknowledgments

We are grateful to all of the similarity detection software vendors and companies that provided us with free access to their systems as well as for the thorough information presented in this survey.

## Limitations

This survey, the information presented and the results are all limited by the plagiarism detection systems to which we had access as well as the vendors that agreed to grant us with a free trial of their respective software.

## References

Ali, A., Abdulla, H., & Snášel, V. (2011). Overview and Comparison of Plagiarism Detection Tools. *VSB-Technical University of Ostrava*.

Birkić, T., Celjak, D., Cundeković, M., & Rako, S. (2016). Analysis of software for plagiarism detection in science and education. *Report. University of Zagreb, 26p*.

Chow, Y., Susilo, W., Pranata, I., Barmawi, A., 2016. Detecting visual spoofing using classical cryptanalysis methods in plagiarism detection systems. undefined.

Chowdhury, H.A., Bhattacharyya, D.K., 2018. Plagiarism: Taxonomy, Tools and Detection Techniques. arXiv. https://arxiv.org/abs/1801.06323

Comas-Forgas, R., & Sureda-Negre, J. (2010). Academic Plagiarism: Explanatory Factors from Students' Perspective. *Journal of Academic Ethics*, *8*(3), 217–232. https://doi.org/10.1007/s10805-010-9121-0

Eassom, H. (2016, February 2). *10 types of plagiarism in research*. https://www.wiley.com/network/researchers/submission-and-navigating-peer-review/10-types-of-plagiarism-in-research

Evans Connor. (2020). *Six Proven ways to cheat Turnitin - Paper Per Hour*. https://paperperhour.com/six-proven-ways-cheat-turnitin/

Fishman, T. (2009). "We know it when we see it" is not good enough: toward a standard definition of plagiarism that transcends theft, fraud, and copyright. *4th Asia Pacific Conference on Educational Integrity (4APCEI)*. https://ro.uow.edu.au/apcei/09/papers/37

Foltýnek, T., Dlabolová, D., Anohina-Naumeca, A., Razı, S., Kravjar, J., Kamzola, L., Guerrero-Dib, J., Çelik, Ö., & Weber-Wulff, D. (2020). Testing of support tools for plagiarism detection. *International Journal of Educational Technology in Higher Education*, *17*(1), 46. https://doi.org/10.1186/s41239-020-00192-4

Foltýnek, T., Meuschke, N., & Gipp, B. (2019). Academic plagiarism detection: A systematic literature review. In *ACM Computing Surveys* (Vol. 52, Issue 6). Association for Computing Machinery. https://doi.org/10.1145/3345317

International Organization for Standardization [ISO]. (2001, June). *ISO/IEC 9126-1:2001 Software engineering — Product quality*. http://www.arisa.se/compendium/node6.html

Jharotia, A.-K., 2018. Plagiarism Detection Through Software in Digital World. JK Business School, Gurugram, Haryana.

Khan, N., Agrawal, C., Nishat Ansari, T., 2018. A Review on Various Plagiarism Detection Systems Based on Exterior and Interior Method. IJARCCE 7, 6–12. https://doi.org/10.17148/ijarcce.2018.792

Kumar, S., Boriwal, C., 2019. Plagiarism Issues: Types, Tools and Remedies, Indian Journal of Agricultural Library and Information Services.

Lim, S. (2009). How and why do college students use Wikipedia? *Journal of the American Society for Information Science and Technology*, *60*(11), 2189–2202. https://doi.org/10.1002/asi.21142

Maurer, H., Kappe, F., & Zaka, B. (2006). Plagiarism - A Survey. *Journal of Universal Computer Science*, *12*, 1050–1084.

Mozgovoy, M., Kakkonen, T., & Cosma, G. (2010). Automatic student plagiarism detection: Future perspectives. *Journal of Educational Computing Research*, *43*(4), 511–531. https://doi.org/10.2190/EC.43.4.e

Nahas, M. N. (2017). Survey and Comparison between Plagiarism Detection Tools. *Mahmoud Nadim Nahas. Survey and Comparison between Plagiarism Detection Tools. American Journal of Data Mining and Knowledge Discovery*, *2*(2), 50–53. https://doi.org/10.11648/j.ajdmkd.20170202.12

Park, C. (2003). In other (People's) words: Plagiarism by university students-literature and lessons. In *Assessment and Evaluation in Higher Education* (Vol. 28, Issue 5, pp. 471–488). Taylor & Francis Group . https://doi.org/10.1080/02602930301677

Selemani, A., Chawinga, W. D., & Dube, G. (2018). Why do postgraduate students commit plagiarism? An empirical study. *International Journal for Educational Integrity*, *14*(1), 7. https://doi.org/10.1007/s40979-018-0029-6

Shkodkina, Y., & Pakauskas, D. (2017). Comparative Analysis of Plagiarism Detection Systems. *Business Ethics and Leadership*, *1*(3), 27–35. https://doi.org/10.21272/bel.1(3).27-35.2017

Smart, P., Gaston, T., 2019. How prevalent are plagiarized submissions? Global survey of editors. Learned Publishing 32, 47–56. https://doi.org/10.1002/leap.1218

Sobhagyawati, G., 2017. Plagiarism Detection Software: an Overview, in: Emerging Trends and Technology in Knowledge Management. Prateeksha Publications, Jaipur, pp. 129–143.

Turnitin. (2012). *The Plagiarism Spectrum (white paper)*. https://www.turnitin.com/resources/plagiarism-spectrum-2-0

Vani, K., Gupta, D., 2016. Study on extrinsic text plagiarism detection techniques and tools. Journal of Engineering Science and Technology Review 9, 150–164. https://doi.org/10.25103/jestr.094.23

Velásquez, J. D., Covacevich, Y., Molina, F., Marrese-Taylor, E., Rodríguez, C., & Bravo-Marquez, F. (2016). DOCODE 3.0 (DOcument COpy DEtector): A system for plagiarism detection by applying an information fusion process from multiple documental data sources. *Information Fusion*, *27*, 64–75. https://doi.org/10.1016/j.inffus.2015.05.006

Weber-Wulff, D. (2014). Plagiarism and Academic Misconduct. In *False Feathers* (pp. 3–27). Springer Berlin Heidelberg. https://doi.org/10.1007/978-3-642-39961-9_2

Weber-Wulff, D. (2015). Plagiarism Detection Software: Promises, Pitfalls, and Practices. In *Handbook of Academic Integrity* (pp. 1–10). Springer Singapore. https://doi.org/10.1007/978-981-287-079-7_19-1